\documentclass[final]{siamltex}
\usepackage[]{amsmath,amssymb,epsfig}

\usepackage{graphicx}
\usepackage{subfig}
\usepackage{algorithm}
\usepackage{algorithmic}

\begin{document}

\title{Uniqueness of Low-Rank Matrix Completion by Rigidity Theory}
\author{
Amit~Singer%
\thanks{Department of Mathematics and PACM, Princeton University, Fine Hall, Washington Road, Princeton NJ 08544-1000 USA, email: amits@math.princeton.edu}
\and Mihai Cucuringu%
\thanks{PACM, Princeton University, Fine Hall, Washington Road, Princeton NJ 08544-1000 USA, email: mcucurin@math.princeton.edu}
}

\date{}
\maketitle

\begin{abstract}
The problem of completing a low-rank matrix from a subset of its entries is often encountered in the analysis of incomplete data sets exhibiting an underlying factor model with applications in collaborative filtering, computer vision and control. Most recent work had been focused on constructing efficient algorithms for exact or approximate recovery of the missing matrix entries and proving lower bounds for the number of known entries that guarantee a successful recovery with high probability. A related problem from both the mathematical and algorithmic point of view is the distance geometry problem of realizing points in a Euclidean space from a given subset of their pairwise distances. Rigidity theory answers basic questions regarding the uniqueness of the realization satisfying a given partial set of distances. We observe that basic ideas and tools of rigidity theory can be adapted to determine uniqueness of low-rank matrix completion, where inner products play the role that distances play in rigidity theory. This observation leads to an efficient randomized algorithm for testing both local and global unique completion. Crucial to our analysis is a new matrix, which we call the {\em completion matrix}, that serves as the analogue of the rigidity matrix.
\end{abstract}



\section{Introduction}

Can the missing entries of an incomplete real valued matrix be recovered? Clearly, a matrix can be completed in an infinite number of ways by replacing the missing entries with arbitrary values. In order for the completion question to be of any value we must restrict the matrix to belong to a certain class of matrices. A popular class of matrices are the matrices of limited rank and the problem of completing a low-rank matrix from a subset of its entries has received a great deal of attention lately. The completion problem comes up naturally in a variety of settings. One of these is the {\em Netflix} problem \cite{Netflix}, where users submit rankings for only a small subset of movies, and one would like to infer their preference of unrated movies. The data matrix of all user-ratings may be approximately low-rank because it is believed that only a few factors contribute to an individual's preferences. The completion problem also arises in computer vision, in the problem of inferring three-dimensional structure from motion \cite{Kanade}, as well as in many other data analysis, machine learning \cite{SrebroPHD}, control \cite{Mesbani} and other problems that are modeled by a factor model. Numerous completion algorithms have been proposed over the years, see e.g., \cite[and references therein]{Jacobs,Srebro2003,Fazel1,FazelPHD,CandesSVT}. Many of the algorithms relax the non-convex rank constraint by the convex set of semidefinite positive matrices and solve a convex optimization problem using semidefinite programming (SDP) \cite{SDP}. Recently, using techniques from compressed sensing \cite{CandesTao1,Donoho}, Cand\`es and Recht \cite{CandesCompletion} proved that if the pattern of missing entries is random then the minimization of the convex nuclear norm (the $\ell_1$ norm of the singular values vector) finds (with high probability) the exact completion of most $n\alpha\times n$ matrices of rank $d$ as long as the number of observed entries $m$ satisfies $m\geq C(\alpha)dn^{1.2}\log n$, where $C(\alpha)$ is some function. Even more recently, Keshavan, Oh, and Montanari \cite{Montanari1,Montanari2} improved the bound to $C(\alpha) dn\log n$ and also provided an efficient completion algorithm.

These fascinating recent results do not provide, however, a solution to the more practical case in which the pattern of missing entries is non-random. Given a specific pattern of missing entries, extremely desirable would be an algorithm that can determine the uniqueness of a rank-$d$ matrix completion. Prior to running any of the numerous existing completion algorithms such as SDP it is important for the analyst to know if such a completion is indeed unique.

Building on ideas from rigidity theory (see, e.g., \cite{Roth1981}) we propose an efficient randomized algorithm that determines whether or not it is possible to uniquely complete an incomplete matrix to a matrix of specified rank $d$. Our proposed algorithm does not attempt to complete the matrix but only determines if a unique completion is possible.
We introduce a new matrix, which we call {\em the completion matrix} that serves as the analogue of the rigidity matrix in rigidity theory. The rank of the completion matrix determines a property which we call infinitesimal completion. Whenever the completion matrix is large and sparse its rank can be efficiently determined using iterative methods such as LSQR \cite{LSQR}. As in rigidity theory, we will also make the distinction between {\em local} completion and {\em global} completion. The analogy between rigidity and completion is quite striking, and we believe that many of the results in rigidity theory can be usefully translated to the completion setup. Our randomized algorithm for testing local completion is based on a similar randomized algorithm for testing local rigidity that was suggested by Hendrickson \cite{Hendrickson1992}, whereas our randomized algorithm for testing global completion is based on the recent randomized global rigidity testing algorithm of Gortler, Healy, and Thurston \cite{Harvard1} who proved a conjecture by Connelly \cite{Connelly} for the characterization of globally rigid frameworks. Due to the large body of existing work in rigidity theory we postpone some of the translation efforts to the future.

The organization of the paper is as follows. Section \ref{sec:rigidity} contains a glossary of definitions and results in rigidity theory on which our algorithms are based. In Section \ref{sec:gram} we analyze the low-rank completion problem for the particular case of positive semidefinite Gram matrices and present algorithms for testing local and global completion of such matrices. In Section \ref{sec:svd} the analysis is generalized to the more common completion problem of general low-rank rectangular matrices and corresponding algorithms are provided. Section \ref{sec:comb} is concerned with the combinatorial characterization of entry patterns that can be either locally completed or globally completed. In particular, we present a simple combinatorial characterization for rank-1 matrices and comment on the rank-2 and rank-$d$ ($d\geq 3$) cases. In Section \ref{sec:numerical} we detail the results of extensive numerical simulations in which we tested the performance of our algorithms while verifying the theoretical bounds of \cite{CandesCompletion,Montanari1,Montanari2} for matrices with random missing patterns. Finally, Section \ref{sec:summary} is a summary and discussion.

\section{Rigidity theory: basic definitions and results}
\label{sec:rigidity}
Rigidity theory tries to answer if a given partial set of distances $d_{ij}=\|p_i-p_j\|$ between $n$ points in $\mathbb{R}^d$ uniquely determines the coordinates of the points $p_1,\ldots,p_n$ up to rigid transformations (translations, rotations, reflections).
This section is a self contained but extremely selective and incomplete collection of basic definitions and results in rigidity theory from the literature (e.g., \cite[and references therein]{Roth1981,Harvard1,Connelly2,Hendrickson1992}). Readers who are unfamiliar with rigidity theory may wish to skip this section at first reading and use it as a glossary.

A {\em bar and joint framework} in $\mathbb{R}^d$ is an undirected graph $G=(V,E)$ ($|V|=n, |E|=m$) and a {\em configuration} $p$
which assigns a point $p_i$ in $\mathbb{R}^d$ to each vertex $i$ of the graph. The edges of the graph correspond to distance constraints, that is, $(i,j)\in E$ iff there is a bar of length $d_{ij}$ between $p_i$ and $p_j$.
Consider a motion of the configuration with $p_i(t)$ being the displacement vector of vertex $i$ at time $t$. Any smooth motion that instantaneously preserves the distance $d_{ij}$ must satisfy $\frac{d}{dt}\|p_i-p_j\|^2 = 0$ for all $(i,j)\in E$. Denoting the instantaneous velocity of the $i$-th point by $\dot{p}_i$, it follows that
\begin{equation}
\label{rigidity-eqs}
(p_i - p_j) \cdot (\dot{p}_i - \dot{p}_j) = 0, \;\mbox{for all } (i,j)\in E.
\end{equation}
Given a framework $G(p)$ in $\mathbb{R}^d$, a solution $\dot{p} = [\dot{p}_1^T \, \dot{p}_2^T \, \cdots \, \dot{p}_n^T]^T$ with $\dot{p}_i$ in $\mathbb{R}^d$ to the system of linear equations (\ref{rigidity-eqs}) is called an {\em infinitesimal motion}.
This linear system consisting of $m$ equations in $dn$ unknowns can be brought together as $R_G(p)\dot{p} = 0$, where $R_G(p)$ is the so called $m\times dn$ {\em rigidity matrix}.

Note that for every skew-symmetric $d\times d$ matrix $A$ (with $A^T=-A$) and for every $b\in \mathbb{R}^d$ we have that $\dot{p}_i = Ap_i + b$ is an infinitesimal motion, where $A$ accounts for some orthogonal transformation and $b$ accounts for some translation. Such infinitesimal motions are called {\em trivial} because they are the derivative of rigid transformations.
A framework $G(p)$ is {\em infinitesimally rigid} if all infinitesimal motions are trivial, and {\em infinitesimally flexible} otherwise. Observe that the trivial infinitesimal motions span a $d(d+1)/2$ dimensional subspace of $\mathbb{R}^{dn}$, combining the $d$ degrees of freedom of translations with the $d(d-1)/2$ degrees of freedom of orthogonal transformations. Therefore, the dimensionality of the null space of the rigidity matrix of a given framework determines if it is infinitesimally rigid or infinitesimally flexible: if $\operatorname{dim} \operatorname{null}(R_G(p)) > d(d+1)/2$ then $G(p)$ is infinitesimally flexible, otherwise $\operatorname{dim} \operatorname{null}(R_G(p)) = d(d+1)/2$ and $G(p)$ is infinitesimally rigid.

A framework $G(p)$ is said to be {\em locally rigid} if there exists a neighborhood $U$ of
$G(p)$ such that $G(p)$ is the only framework in $U$ with the same set of edge lengths, up to
rigid transformations. In other words, there is no continuous deformation that preserves the edge lengths.

A configuration is {\em generic} if the coordinates do not satisfy any non-zero
polynomial equation with integer coefficients (or equivalently algebraic coefficients). Generic configurations are not
an open set of configurations in $\mathbb{R}^{dn}$, but they do form a dense set of full
measure.

The following rigidity predictor was introduced by Gluck \cite{Gluck} and extensively
used in Asimow and Roth \cite{Asimow}:
\begin{theorem}
[Gluck \cite{Gluck}, Asimow and Roth \cite{Asimow}]\label{Th-Asimow} If a generic framework $G(p)$ of a graph $G$ with $d+1$
or more vertices is locally rigid in $\mathbb{R}^d$, then it is infinitesimally rigid; otherwise $\operatorname{dim} \operatorname{null}(R_G(p))>d(d+1)/2$.
\end{theorem}

Since the dimension of the null space of the rigidity matrix is the same at every
generic point, local rigidity in $\mathbb{R}^d$ is a generic property.
That is, either all generic frameworks of the graph $G$
are locally rigid, or none of them are. This is a condition for {\em generic local
rigidity} in $\mathbb{R}^d$ which can be considered as a property of the graph $G$.

Hendrickson  \cite{Hendrickson1992} observed that generic local rigidity can therefore be tested efficiently in any dimension using a randomized algorithm: simply randomize the displacement vectors $p_1,\ldots,p_n$ while ignoring the specific distance constraints that they have to satisfy, construct the rigidity matrix corresponding to the framework of the original graph with the randomized points and check its rank. With probability one, the rank of the rigidity matrix that corresponds to the unknown true displacement vectors equals the rank of the randomized rigidity matrix. A similar randomized algorithm for local generic rigidity was described in \cite[Algorithm 3.2]{Harvard1}.

Since generic local rigidity is a combinatorial property of the graph, it is natural to search for a combinatorial characterization of such graphs. Such a combinatorial characterization exists for rigidity in the plane ($d=2$).
The total number of degrees of freedom for $n$ points in the plane is $2n$. How many distance constraints are necessary to limit a framework to having only the trivial motions? Or equivalently, how many edges are necessary for a graph to be rigid? Each edge can remove a single degree of freedom. Rotations and translations will always be possible, so at least $2n-3$ edges are necessary for a graph to be rigid. For example, a graph with $n=3$ and $|E|=2n-3=3$ edges is the triangle which is rigid. Similarly, a graph with $n = 4$  and $|E|=2n-3 = 5$ is $K_4$ minus one edge which is also locally rigid. However, the graph on $n = 5$ vertices consisting of $K_4$ plus one dangling node has $2n-3 = 7$ edges but it is not rigid. Such edge counting considerations had already been made by Maxwell \cite{Maxwell} in the 19th century. Laman \cite{Laman} was the first to prove the precise combinatorial characterization of rigid frameworks in the plane:

A framework is {\em minimally rigid}, if it is infinitesimally flexible once an edge is removed. A framework is {\em redundantly rigid}, if it is infinitesimally rigid upon the removal of any single edge.
\begin{theorem}[Laman \cite{Laman}]\label{Th-Laman}
A graph with $n$ vertices is generically minimally rigid in 2D if and only if it has $2n-3$ edges and no subgraph of $n'$ vertices has more than $2n'-3$ edges. A graph is generically rigid if it contains a Laman graph with $n$ vertices.
\end{theorem}

In other words, Laman condition for minimally rigid graphs says that the graph need to have at least $2n-3$ ``well-distributed'' edges. Generic rigidity is a property of the graph connectivity, not the geometry. The {\em pebble game} algorithm of Hendrickson and Jacobs \cite{Pebble1997} applies Laman's theorem to determine generic local rigidity of a given graph in at most $O(n^2)$ steps. Laman graphs (i.e., generic minimally rigid graphs) are an instance of tight sparse graphs. A graph with $n$ vertices and $m$ edges is said to be {\em $(k,l)$-sparse} if every subset of $n'\leq n$ vertices spans at most $kn'-l$ edges. If,
furthermore, $m = kn-l$, the graph is called {\em tight} (see, e.g, \cite{Lee}). Thus the $(2,3)$-sparse tight graphs are the Laman graphs.
Unfortunately, an exact combinatorial characterization of locally generic rigid graphs is currently not available in higher dimensions ($d\geq 3$). The ``edge-counting" condition is necessary but not sufficient and it is a long standing open problem what is the combinatorial condition for rigidity in 3D.

Local generic rigidity does not imply unique realization of the framework. For example, consider the $2D$-rigid graph with $n=4$ vertices and $m=5$ edges consisting of two triangles that can be folded with respect to their joint edge.
A framework $G(p)$ is {\em globally rigid} in $\mathbb{R}^d$ if all frameworks $G(q)$ in $\mathbb{R}^d$
which are $G(p)$-equivalent (have all bars the same length as $G(p)$) are congruent
to $G(p)$ (that is, they are related by a rigid transformation).

Hendrickson proved two key necessary conditions for global rigidity of a framework at a generic configuration:
\begin{theorem}[Hendrickson \cite{Hendrickson1992}] \label{Th-Hendrickson}
If a framework $G(p)$, other than a simplex, is globally rigid for a generic configuration $p$ in $\mathbb{R}^d$ then:
\begin{itemize}
\item The graph $G$ is vertex $(d + 1)$-connected;
\item The framework $G(p)$ is redundantly rigid, in the sense that removing any one edge leaves a graph which is infinitesimally rigid.
\end{itemize}
\end{theorem}

A graph $G$ is {\em generically globally rigid} in $\mathbb{R}^d$ if $G(p)$ is globally rigid at
all generic configurations $p$ \cite{Connelly3,Connelly}. Only recently it was demonstrated that global rigidity is
a generic property in this sense for graphs in each dimension \cite{Connelly,Harvard1}.   The conditions of Hendrickson as stated in Theorem \ref{Th-Hendrickson} are necessary for generic global
rigidity. They are also sufficient on the line, and in the plane \cite{JacksonJordan}. However,
by a result of Connelly \cite{Connelly3}, $K_{5,5}$ in 3-space is generically redundantly rigid
and 5-connected but is not generically globally rigid.

The critical technique used for proving global rigidity of frameworks uses stress
matrices. A {\em stress} is an assignment of scalars $w_{ij}$ to the edges
such that for each $i\in V$
\begin{equation}
\sum_{j:\,(i,j)\in E} \omega_{ij} (p_i-p_j) = 0.
\end{equation}
Alternatively, a self-stress is a vector in the left null space of the rigidity matrix: $R_G(p)^T w = 0$.
A stress vector can be rearranged into an $n\times n$ symmetric matrix $\Omega$, known as the {\em stress matrix}, such that for $i\neq j$, the $(i,j)$ entry of $\Omega$ is $\Omega_{ij}=-\omega_{ij}$, and the diagonal entries for $(i,i)$ are $\Omega_{ii} = \sum_{j:\, j\neq i} \omega_{ij}$.
Note that all row and column sums are now zero from which it follows that the all-ones vector $(1 \; 1 \; \cdots \; 1)^T$ is in the null space of $\Omega$ as well as each of the coordinate vectors of the configuration $p$. Therefore, for generic configurations the rank of the stress matrix is at most $n-(d+1)$.
The key connection for global rigidity are the following pair of results:
\begin{theorem}[Connelly \cite{Connelly}] \label{Th-Connelly}
If $p$ is a generic configuration in $\mathbb{R}^d$, such that
there is an equilibrium stress, where the rank of the associated stress matrix
$\Omega$ is $n-(d+1)$, then $G(p)$ is globally rigid in $\mathbb{R}^d$.
\end{theorem}

\begin{theorem}
[Gortler, Healy, and Thurston \cite{Harvard1}] \label{Th-Harvard}
Suppose that $p$ is a generic configuration in $\mathbb{R}^d$, such that $G(p)$ is globally rigid in $\mathbb{R}^d$. Then either $G(p)$
is a simplex or there is an equilibrium stress where the rank of the associated
stress matrix $\Omega$ is $n-(d+1)$.
\end{theorem}

Based on their theorem, Gortler, Healy and Thurston also provided a randomized polynomial algorithm for checking generic global rigidity of a graph \cite[Algorithm 3.3]{Harvard1}. If the graph is generically local rigid then their algorithm picks a random stress vector of the left null space of the rigidity matrix, converts it into a stress matrix and computes the rank of the stress matrix which is compared with $n-(d+1)$ to determine generic global rigidity.

Rigidity properties of random Erd\H{o}s-R\'enyi $G(n, \beta)$ graphs where each edge is chosen with probability $\beta$ have been recently analyzed for the two dimensional case:
\begin{theorem}[Jackson, Servatius, and Servatius \cite{Jackson}]\label{Th-Jackson}
Let $G=G(n, \beta)$, where $\beta = (\log n + k \log \log n + w(n))/n$, and
$\lim_{n\to \infty} w(n)=+\infty$.
\begin{itemize}
\item If $k = 2$ then G is asymptotically almost surely (a.a.s) generically locally rigid.
\item If $k = 3$ then G is a.a.s. generically globally rigid.
\end{itemize}
\end{theorem}
The bounds on $p$ given in Theorem \ref{Th-Jackson} are best possible in the sense that $G=G(n, \beta)$
and $\beta = (\log n + k \log \log n + c)/n$ for any constant $c$, then $G$ does not a.a.s. have
minimum degree at least $k$. The emergence of large rigid components was also studied by Theran in \cite{Theran}.

Supplied with this rigidity theory background, we are ready to analyze the low-rank matrix completion problem.
\section{Gram matrices}
\label{sec:gram}
We start by analyzing the completion problem of low-rank positive semidefinite Gram matrices with missing entries. We make extensive use of the terminology and results summarized in the glossary Section \ref{sec:rigidity} which the reader is advised to consult whenever felt needed.

For a collection of $n$ vectors $p_1,p_2,\ldots,p_n \in \mathbb{R}^d$ there corresponds an $n\times n$ Gram matrix $J$ of rank $d$ whose entries are given by the inner products
\begin{equation}
J_{ij} = p_i \cdot p_j ,\quad i,j=1,\ldots,n. \label{Gram}
\end{equation}
An alternative way of writing $J$ is through its Cholesky decomposition $J = P^T P$,
where $P$ is a $d\times n$ matrix given by
$P = \left[
             \begin{array}{cccc}
               p_1 & p_2 & \cdots & p_n \\
             \end{array}
           \right]$,
from which it is clear that $\operatorname{rank}(J) = d$.
If $J$ is fully observed (no missing entries) then the Cholesky decomposition of $J$ reveals $P$ up to a $d\times d$ orthogonal matrix $O$ $(OO^T=I)$, as $J=P^T P = (OP)^T (OP)$. Now, suppose that only a few of the entries of $J$ are observed by the analyst. The symmetry of the matrix $J$ implies that the set of observed entries defines an undirected graph $G=(V,E)$ with $n$ vertices where $(i,j)\in E$ is an edge iff the entry $J_{ij}$ is observed. The graph may include self loop edges of the form $(i,i)$ corresponding to observed diagonal elements $J_{ii}$ (note that rigidity graphs have no self loops as the distance from a point to itself is not informative). For an incomplete Gram matrix $J$ with an observed pattern that is given by the graph $G$, we would like to know if it is possible to {\em uniquely} complete the missing entries of $J$ so that the resulting completed matrix is of rank $d$.

For example, consider the three planar points
\begin{equation}
\label{ex-R1}
p_1 = \left(
        \begin{array}{c}
          0 \\
          1 \\
        \end{array}
      \right),\;
p_2 = \left(
        \begin{array}{c}
          1 \\
          2 \\
        \end{array}
      \right),\;
p_3 = \left(
        \begin{array}{c}
          2 \\
          3 \\
        \end{array}
      \right): \; P = \left(
      \begin{array}{ccc}
        0 & 1 & 2 \\
        1 & 2 & 3 \\
      \end{array}
    \right),
\end{equation}
whose corresponding 3-by-3 Gram matrix $J$ is of rank 2
\begin{equation}
J = P^T P = \left(
      \begin{array}{ccc}
        1 & 2 & 3 \\
        2 & 5 & 8 \\
        3 & 8 & 13 \\
      \end{array}
    \right).
\end{equation}
The following three different missing patterns demonstrate that a matrix may either have a unique completion, a finite number of possible completions, or an infinite number of possible completions:
\begin{equation}
\label{ex-G1}
\left(
      \begin{array}{ccc}
        1 & 2 & 3 \\
        2 & 5 & 8 \\
        3 & 8 & ? \\
      \end{array}
    \right),\quad
    \left(
      \begin{array}{ccc}
        1 & 2 & ? \\
        2 & 5 & 8 \\
        ? & 8 & 13 \\
      \end{array}
    \right),\quad
\left(
      \begin{array}{ccc}
        1 & 2 & 3 \\
        2 & ? & 8 \\
        3 & 8 & ? \\
      \end{array}
    \right).
\end{equation}
For the left matrix, the missing diagonal element is uniquely determined by the fact that $\det J = 0$. For the middle matrix, the vanishing determinant constraint is a quadratic equation in the missing entry and there are {\em two} different possible completions (the reader may check that $J_{13}=J_{31}=3.4$ is a valid completion). For the right matrix, the vanishing determinant constraint is a single equation for the two unknown diagonal elements, and so there are infinite number of possible completions. We want to go beyond 3-by-3 matrices and develop techniques for the analysis of much larger matrices with arbitrary patterns of missing entries.

\subsection{The completion matrix and local completion}
We now adopt rigidity theory to the matrix completion problem. We start by considering motions $p_i(t)$ that preserve the inner products $J_{ij} = p_i \cdot p_j $ for all $(i,j)\in E$. Differentiating (\ref{Gram}) with respect to $t$ yields the set of $m=|E|$ linear equations for the unknown velocities
\begin{equation}
\label{completion}
p_i \cdot \dot{p}_j + p_j \cdot \dot{p}_i = 0,\quad \mbox{for all } (i,j)\in E.
\end{equation}
This linear system can be brought together as $C_G(p)\dot{p} = 0$, where
$C_G(p)$ is an $m\times dn$ coefficient matrix which we call the {\em completion matrix}. The completion matrix is sparse as it has only $2d$ non-zero elements per row. The locations of the non-zero entries are solely determined by the graph $G$, while their values are determined by the particular realization $p$. The solution space of (\ref{completion}) is at least $d(d-1)/2$ dimensional, due to orthogonal transformations that preserve inner products. Indeed, substituting into (\ref{completion}) the ansatz $\dot{p_i} = Ap_i$ (for all $i=1,\ldots,n$) with $A$ being a constant $d\times d$ matrix yields $p_i^T (A+A^T) p_j = 0$. Therefore, any choice of a skew-symmetric matrix $A=-A^T$ leads to a possible solution. We refer to these as the trivial infinitesimal motions. The number of trivial degrees of freedom in the completion problem is $d(d-1)/2$ which differs from its rigidity theory counterpart, because translations preserve distances but not inner products.

The rank of the completion matrix will help us determine if a unique completion of the Gram matrix $J$ is possible. Indeed, if
\begin{equation}
\label{cond}
\operatorname{dim} \operatorname{null}(C_G(p)) > d(d-1)/2,
\end{equation}
that is, if the dimensionality of the null space of $C_G(p)$ is greater than $d(d-1)/2$, then there exist non-trivial solutions to (\ref{completion}). In other words, there are non-trivial infinitesimal flexible motions that preserve the inner products. The rank of the completion matrix thus determines if the framework is {\em infinitesimally completable}. The counterpart of Asimow-Roth theorem would imply that there exists a non-trivial transformation that preserves the inner products and that the matrix is not {\em generically locally completable}. Since almost all completion matrices for a given graph have the same rank, generic local completion is a property of the graph itself, and we do not need any advance knowledge of the realization of the matrix that we are trying to complete. Instead, we can simply construct a completion matrix from a randomly chosen realization. With probability one, the dimensionality of the null space of the randomized completion matrix will be the same as that of the completion matrix of the true realization, and this rank will determine if the Gram matrix is generically locally completable or not. The resulting randomized algorithm for testing local completion is along the same lines of the randomized algorithms for testing local rigidity \cite{Hendrickson1992,Harvard1}:
\newline
\begin{algorithm}[h]
\caption{Local completion of $n\times n$ rank-$d$ Gram matrices}\label{alg:local-gram}
\begin{algorithmic}[1]
\REQUIRE Graph $G=(V,E)$ with $n$ vertices and $m$ edges corresponding to known matrix entries (self loops are possible).
\STATE Randomize a realization $p_1,\ldots,p_n$ in $\mathbb{R}^d$.
\STATE Construct the sparse completion matrix $C_G(p)$ of size $m\times dn$.
\STATE Check if there is a non-trivial infinitesimal motion $\dot{p}$ satisfying $C_G(p)\dot{p}=0$.\label{alg-rank}
\STATE If a non-trivial infinitesimal motion exists then $G$ cannot be locally completable, otherwise $G$ is locally completable.
\end{algorithmic}
\end{algorithm}

In order for Algorithm \ref{alg:local-gram} to be feasible for large-scale problems, the approach sketched above requires a fast method to determine the existence of a non-trivial infinitesimal motion, which is equivalent to checking that the rank of the completion matrix satisfies $\operatorname{rank}(C_G(p)) < dn-d(d-1)/2$. This is not a straightforward check from the numerical linear algebra point of view. Note that the rank of a matrix can be determined only up to some numerical tolerance (like machine precision), because the matrix may have arbitrarily small non-zero singular values. The full singular value decomposition (SVD) is the most reliable way to compute the rank of the completion matrix (e.g., using MATLAB's \textsf{rank} function), but is also the most time consuming and is computationally prohibitive for large matrices. The completion matrix is often sparse, in which case, sparse LU, sparse QR and rank-revealing factorizations \cite[and references within]{GotsmanToledo,TimDavis} are much more efficient. However, due to non-zero fill-ins, such methods quickly run out of memory for large scale problems. Our numerical experimentation with Gotsman and Toledo's sparse LU  MATLAB function \textsf{nulls} \cite{GotsmanToledo} and Davis' SuiteSparseQR MATLAB library and \textsf{spqr} function \cite{TimDavis} encountered some memory issues for $n\geq 6000$. Therefore, for large-scale sparse completion matrices we use iterative methods that converge fast and do not have special storage requirements.
The iterative procedure has several steps:
\begin{enumerate}
\item Use different choices of skew-symmetric $d\times d$ matrices $A$ to construct $d(d-1)/2$ linearly independent trivial infinitesimal motions (recall that trivial motions are given by $\dot{p}_i = Ap_i$, $i=1,\ldots,n$), and store the trivial motions in a $dn\times d(d-1)/2$ matrix $T$.
\item Compute the QR factorization of $T=QR$ (see, e.g., \cite{Golub}) such that the columns of the $dn\times d(d-1)/2$ matrix $Q$ form an orthogonal basis for the subspace of trivial motions, i.e.,  the two column spaces are the same $\operatorname{col}(Q)=\operatorname{col}(T)$ and $Q^T Q = I_{d(d-1)/2}$.
\item Randomize a unit size vector $b\in \mathbb{R}^{dn}$ in the orthogonal subspace of trivial motions, $b \in \operatorname{col}(Q)^\perp$. This is performed by randomizing a vector $v\in \mathbb{R}^{dn}$ with i.i.d standard Gaussian entries ($v_i \sim \mathcal{N}(0,1)$), projecting $v$ onto the orthogonal subspace using $w=(I-QQ^T)v$ (the matrix $QQ^T$ is never formed), and normalizing $b = w / \|w\|$. It is easy to check that $Q^T b = 0$ and that $b$ has the desired normal distribution.
\item Attempt solving the linear system $C_G(p)^T x=b$ for the unknown $x$ using an iterative method such as LSQR \cite{LSQR} that minimizes the sum of squares residual. The linear system may or may not have a solution. Numerically, a tolerance parameter $tol$ must be supplied to the LSQR procedure. Set the tolerance parameter to be $tol = \varepsilon n^{-1/2}$, with a small $\varepsilon$, e.g., $\varepsilon=10^{-4}$. If the residual cannot be made smaller then $tol$ then conclude that the linear system has no solution. In such a case $b \not \in \operatorname{col}(C_G(p)^T)$ and by the fundamental theorem of linear algebra it follows that the projection of $b$ onto $\operatorname{null}(C_G(p))$ is non-zero. Since $b$ is orthogonal to all trivial motions, it follows that $\operatorname{null}(C_G(p))$ contains a non-trivial infinitesimal motion and the matrix is not locally completable. On the other hand, if the linear system $C_G(p)x=b$ has a solution in the sense that the residual is smaller than $\varepsilon n^{-1/2}$ then the projection of $b$ on $\operatorname{null}(C_G(p))$ is smaller than $\varepsilon n^{-1/2}$. If a non-trivial infinitesimal motion $\dot{p}$ exists, then the projection of $b$ onto it is normally distributed with zero mean and variance $n^{-1}$, that is $b^T \dot{p}/\|\dot{p}\| \sim \mathcal{N}(0,n^{-1})$.  Therefore, with probability at least $1-\displaystyle{\frac{2\varepsilon}{\sqrt{2\pi}}}$ all infinitesimal motions are trivial.
\end{enumerate}
The LSQR procedure consists of applying the sparse completion matrix $C_G$ and its transpose $C_G^T$ to vectors with no special need for storage. The number of non-zero entries in the completion matrix is $2dm$ which is also the computational cost of applying it to vectors. The number of LSQR iterations depends on the non-zero singular values and in particular the ratio of the largest singular value and the smallest non-zero singular value (condition number). Arbitrarily small singular values may cause our iterative algorithm with its preset tolerance to fail. In practice, however, at least for moderate values of $n$, the full SVD revealed that such small singular values are rare.

We conclude this section by general remarks on the iterative procedure described above. First, note that the same iterative procedure can be used to determine local rigidity of bar and joint frameworks, and perhaps it can also be useful in other applications where existence of non-trivial null space vector is sought to be determined. Second, iterative methods can be often accelerated by a proper choice of a preconditioner matrix. This leads to the interesting question of designing a suitable preconditioner for completion and rigidity matrices, which we defer for future investigation.

\subsection{Global completion and stress matrices}
Generically local completion of a framework means that the realization cannot be continuously deformed while satisfying the inner product constraints. However, as the middle matrix in example (\ref{ex-G1}) shows, local completion does not exclude the possibility of having a non-trivial discontinuous deformation that satisfies the inner product constraints, where by non-trivial we mean that the deformation is not an orthogonal transformation. We say that the framework is {\em globally completable} if the only deformations that preserve the inner products are the trivial orthogonal transformations (rotations and reflections). While local completion allows for a finite number of different completions, global completion is a stronger property that certifies that completion is unique.

A {\em completion stress} $\omega$ for a framework is an assignment of weights $\omega_{ij}$ on the edges of the graph such that for every vertex $i \in V$
\begin{equation}
\label{completion-stress}
\sum_{j:\, (i,j)\in E} \omega_{ij} p_j = 0.
\end{equation}
Equivalently, a completion stress $\omega$ is a vector in the left null space of the completion matrix $C_G(p)$, i.e., $C_G(p)^T\omega = 0$. A stress matrix $\Omega$ is a symmetric $n\times n$ matrix obtained by the following rearrangement of the completion stress vector entries: $\Omega_{ij} = w_{ij}$ for $(i,j)\in E$, and
$\Omega_{ij} = 0$ for $(i,j)\not \in E$.

It follows from (\ref{completion-stress}) that if $\omega$ is a stress for the framework of $p_1,\ldots,p_n$ then it is also a stress for the framework of $Ap_1,\ldots,Ap_n$, where $A$ is any $d\times d$ linear transformation. In other words, the $d$ coordinate vectors and their linear combinations are in the null space of the stress matrix $\Omega$ and $\operatorname{dim} \operatorname{null}(\Omega) \geq d$. Gortler, Healy, and Thurston \cite{Harvard1} suggested a randomized algorithm for testing global rigidity based on Theorems \ref{Th-Connelly} and \ref{Th-Harvard} that relate stress matrices and global rigidity. Building on these Theorems and their randomized algorithm, we propose Algorithm {\ref{alg:global-gram}, which is a randomized algorithm for testing global completion. Algorithm \ref{alg:global-gram} uses iterative methods in order to insure its scalability to large scale matrices.
\begin{algorithm}[h]
\caption{Global completion of $n\times n$ rank-$d$ Gram matrices}\label{alg:global-gram}
\begin{algorithmic}[1]
\REQUIRE Graph $G=(V,E)$ with $n$ vertices and $m$ edges corresponding to known matrix entries (self loops are possible).
\STATE Check local completion using Algorithm \ref{alg:local-gram}. Proceed only if framework is locally completable.
\STATE Randomize a realization $p_1,\ldots,p_n$ in $\mathbb{R}^d$.
\STATE Construct the sparse completion matrix $C_G(p)$ of size $m\times dn$.
\STATE Compute a random completion stress vector $\omega$ in the left null space of $C_G(p)$ satisfying $C_G(p)^T \omega = 0$.\label{random-stress}
\STATE Rearrange $\omega$ into a completion stress matrix $\Omega$.
\STATE Check if the null space of $\Omega$ contains vectors which are not linear combinations of the $d$ coordinate vectors.\label{rank-stress}
\STATE If such null space vectors exist (i.e., if $\operatorname{dim} \operatorname{null}(\Omega) > d$) then $G$ is not globally completable, otherwise $G$ is globally completable.
\end{algorithmic}
\end{algorithm}

In stage (\ref{random-stress}) of Algorithm \ref{alg:global-gram} we use LSQR, which when initialized with a random starting vector converges to a random left null space solution, rather than to the zero vector. Here we use LSQR with a very small tolerance as we are guaranteed the existence of a non-trivial stress.

In stage (\ref{rank-stress}) we again apply LSQR, this time similarly to the way it is applied in the local completion case (Algorithm \ref{alg:local-gram}). Specifically, we first find a random vector $b$ which is perpendicular to the subspace of coordinate vectors and then we try to solve $\Omega x = b$. Here $\Omega$ is symmetric (compared to $C_G$). For moderate scale problems (e.g., $n\leq 5000$) we use sparse QR (SPQR) to compute the rank of $\Omega$ as it runs faster than LSQR, but cannot handle much larger values of $n$ due to memory problems (fill-ins).

\section{General rectangular low rank matrices}
\label{sec:svd}
A general $n_1\times n_2$ rank-$d$ matrix $X$ can be written as $X=UV^T$ where $U$ is $n_1\times d$ and $V$ is $n_2\times d$ given by $U = \left[
             \begin{array}{cccc}
               u_1 & u_2 & \cdots & u_{n_1} \\
             \end{array}
           \right]^T$ and $V = \left[
             \begin{array}{cccc}
               v_1 & v_2 & \cdots & v_{n_2} \\
             \end{array}
           \right]^T$,
where the $n_1+n_2$ vectors $u_1,\ldots,u_{n_1},v_1,\ldots,v_{n_2}$ are in $\mathbb{R}^d$. Entries of $X$ are inner products of these vectors
\begin{equation}
\label{svd-inner}
X_{ij} = u_i\cdot v_j.
\end{equation}
The observed entries $X_{ij}$ define a bipartite graph $G=(V,E)$ with $n_1+n_2$ vertices as we never observe inner products of the form $u_i \cdot u_{i'}$ or $v_j \cdot v_{j'}$.
Differentiation of (\ref{svd-inner}) with respect to $t$ yields the set of linear equations
\begin{equation}
\label{svd-eqs}
u_i \cdot \dot{v}_j + v_j \cdot \dot{u}_i  = 0, \quad (i,j)\in E
\end{equation}
for the unknown velocities $\dot{u}_i$ and $\dot{v}_j$. The corresponding completion matrix $C_G(u,v)$ has $m$ rows and $d(n_1+n_2)$ columns, but only $2d$ non-zero elements per row.

The decomposition $X=UV^T$ is not unique: if $W$ is an invertible $d\times d$  matrix then $X=U_{W}V_{W}^T$ with $U_W = UW$ and $V_W=V(W^{-1})^T$. It follows that there are $d^2$ degrees of freedom in choosing $W$, because the general linear group $GL(d,\mathbb{R})$ of $d\times d$ invertible matrices is a Lie group over $\mathbb{R}$ of dimension $d^2$. Indeed, substituting in (\ref{svd-eqs}) the ansatz $\dot{u}_i = Au_i$ (for $i=1,\ldots,n_1$) and $\dot{v}_j=Bv_j$ (for $j=1,\ldots,n_2$) gives $u_i^T (B+A^T) v_j = 0$, which is trivially satisfied whenever $B=-A^T$. These are the trivial infinitesimal motions and they span a subspace of dimension $d^2$.

Similarly to local rigidity and local completion of Gram matrices, a condition for local completion for general rectangular matrices is $\operatorname{dim}\operatorname{null}(C_G(u,v)) = d^2$. If the dimension of the null space of the completion matrix is greater than $d^2$ then there are non-trivial deformations that preserve all observed elements of the matrix, where a trivial deformation is any invertible linear transformation of $\mathbb{R}^d$.
The local completion testing algorithm for general rectangular matrices is given in Algorithm \ref{alg:local-svd}.
\begin{algorithm}[h]
\caption{Local completion of $n_1\times n_2$ rank-$d$ matrices}\label{alg:local-svd}
\begin{algorithmic}[1]
\REQUIRE Bipartite graph $G=(V,E)$ with $n_1+n_2$ vertices and $m$ edges corresponding to known matrix entries.
\STATE Randomize a realization $u_1,\ldots,u_{n_1}$ and $v_1,\ldots,v_{n_2}$ in $\mathbb{R}^d$.
\STATE Construct the sparse completion matrix $C_G(u,v)$ of size $m\times d(n_1+n_2)$.
\STATE Check if there is a non-trivial infinitesimal motion $\dot{p}$ satisfying $C_G(p)\dot{p}=0$, where there are $d^2$ trivial motions.
\STATE If a non-trivial infinitesimal motion exists then $G$ cannot be locally completable, otherwise $G$ is locally completable.
\end{algorithmic}
\end{algorithm}

The implementation of the steps in Algorithm \ref{alg:local-svd} is very similar to the implementation of the steps in Algorithm \ref{alg:local-gram} for testing local completion of Gram matrices.

As in the Gram case, local completion does not imply global completion. A stress $\omega$ for the general rectangular matrix case is an assignment of weights $\omega_{ij}$ on the edges of the bipartite graph that satisfy
\begin{equation}
\sum_{j:\,(i,j)\in E} \omega_{ij}v_j = 0, \quad \mbox{for all } i=1,\ldots,n_1,
\end{equation}
and
\begin{equation}
\sum_{i:\,(i,j)\in E} \omega_{ij}u_i = 0, \quad \mbox{for all } j=1,\ldots,n_2.
\end{equation}
The stress $\omega$ is the left null space of the completion matrix, i.e., $C_G(u,v)^T \omega = 0$.

The stress weights can be viewed as the entries of an $n_1\times n_2$ matrix $\tilde{\Omega}$ that satisfies $\tilde{\Omega} V = 0$ and $\tilde{\Omega}^T U = 0$. The stress matrix is the $(n_1+n_2)\times (n_1+n_2)$ symmetric matrix $\Omega$ defined via
\begin{equation}
\Omega = \left(
                                                                                       \begin{array}{cc}
                                                                                         0 & \tilde{\Omega} \\
                                                                                         \tilde{\Omega}^T & 0 \\
                                                                                       \end{array}
                                                                                     \right).
\end{equation}
The stress matrix $\Omega$ satisfies
\begin{equation}
\Omega \left(
         \begin{array}{cc}
           U & 0 \\
           0 & V \\
         \end{array}
       \right) = 0,
\end{equation}
from which it follows that $\operatorname{dim} \operatorname{null}(\Omega) \geq 2d$. Following the randomized algorithms for global rigidity and global completion of Gram matrices, Algorithm \ref{alg:global-svd} is a randomized algorithm for testing global completion of rectangular matrices.
\begin{algorithm}[h]
\caption{Global completion of $n_1 \times n_2$ rank-$d$ matrices}\label{alg:global-svd}
\begin{algorithmic}[1]
\REQUIRE Bipartite graph $G=(V,E)$ with $n_1+n_2$ vertices and $m$ edges corresponding to known matrix entries.
\STATE Check local completion using Algorithm \ref{alg:local-svd}. Proceed only if framework is locally completable.
\STATE Randomize a realization $u_1,\ldots,u_{n_1}$ and $v_1,\ldots,v_{n_2}$ in $\mathbb{R}^d$.
\STATE Construct the sparse completion matrix $C_G(u,v)$ of size $m\times d(n_1+n_2)$.
\STATE Compute a random completion stress vector $\omega$ in the left null space of $C_G(u,v)$ satisfying $C_G(u,v)^T \omega = 0$.
\STATE Rearrange $\omega$ into an $(n_1+n_2)\times (n_1+n_2)$ symmetric completion stress matrix $\Omega$.
\STATE Check if the null space of $\Omega$ contains a vector outside the column space of $\left(
         \begin{array}{cc}
           U & 0 \\
           0 & V \\
         \end{array}\right)$ (equivalently, check if the null space of $\Omega$ contains more than $2d$ linearly independent vectors).
\STATE If $\operatorname{dim} \operatorname{null}(\Omega)>2d$ then $G$ is not globally completable, otherwise $G$ is globally completable.
\end{algorithmic}
\end{algorithm}

\section{Combinatorial approach for local and global completion}
\label{sec:comb}
In the previous sections we observed that the rank of the completion and the stress matrices can determine generic local and global rank-$d$ completion properties of a given graph. These observations led to practical algorithms for property testing, but perhaps there is a simple combinatorial characterization of locally and/or globally completable graphs? In rigidity theory, locally and globally rigid graphs have a simple combinatorial characterization in one and two dimensions (see Theorems \ref{Th-Laman} and \ref{Th-Hendrickson} by Laman and Hendrickson). In higher dimensions, however, such combinatorial characterizations are still missing, and only necessary conditions are available. We therefore believe that exact combinatorial characterization of local and global completion of rank-$d$ matrices for $d\geq 3$ is currently out of reach, and focus our attention to rank-1 and rank-2 matrices and to finding only necessary conditions for $d\geq 3$. We begin by observing some properties of locally completable graphs.

\begin{proposition} \label{gramlam5.5}
In any Gram locally completable rank-$d$ matrix $J$ of size $n \times n$ with $n\geq d$, one can delete entries until the resulting matrix is locally completable and has exactly $dn-\frac{d(d-1)}{2}$ entries.
\end{proposition}

\begin{proof}
The completion matrix of the system of equations (\ref{completion}) $p_i \cdot \dot{p}_j + p_j \cdot \dot{p}_i = 0$ for all  $(i,j)\in E$ must have rank $dn-\frac{d(d-1)}{2}$, and thus one can drop linearly dependent equations until only $dn-\frac{d(d-1)}{2}$ equations remain.
\end{proof}

\begin{proposition} \label{prop_gram_loc}
Any Gram locally completable rank-$d$ matrix $J$ with $|E(G_J)| = dn-\frac{d(d-1)}{2}$ has the property that for any $n' \times n'$ submatrix $J'$ of $J$,  where $d\leq n' \leq n$, it holds true that $|E(G_{J'})| \leq dn'-\frac{d(d-1)}{2}$.
\end{proposition}
\begin{proof}
  For every $(i,j) \in E(G_J)$ there corresponds an equation $p_i \cdot \dot{p}_j + p_j \cdot \dot{p}_i = 0$. Given our interpretation of local completion and Proposition ~\ref{gramlam5.5}, these equations are linearly independent. If for some $n' \times n'$ submatrix $J'$ of $J$, with $n'\geq d+1$ we would have that the number of edges $|E(G_{J'})| > dn'-\frac{d(d-1)}{2} $, then by the same Proposition ~\ref{gramlam5.5} there would be dependence among the corresponding $dn'-\frac{d(d-1)}{2}$ equations, which is a contradiction.
\end{proof}

The result above implies that every locally completable Gram matrix contains an underlying graph that spans all vertices which is $(d,\frac{d(d-1)}{2})$-tight sparse.
It is natural to ask whether any $(d,\frac{d(d-1)}{2})$-tight sparse graph is Gram locally completable. The answer to this question turns out to be false even in one dimension, that is, not every $(1,0)$-tight sparse graph is Gram locally completable for $d=1$.
\begin{proposition}\label{prop:local-gram}
A graph $G$ is minimally Gram locally completable in one dimension if and only if each connected component of $G$ is a tree plus one edge that contains an odd cycle.
\end{proposition}

\begin{proof}
First, suppose $G=(V,E)$ is minimally Gram locally completable for $d=1$. Proposition \ref{gramlam5.5} implies that $G$ has no redundant edges so $|E(G)|=|V(G)|$. Moreover, Proposition \ref{prop_gram_loc} implies that each connected component $H$ of $G$ satisfies $|E(H)|=|V(H)|$, from which it follows that each connected component is a tree plus one edge. We show that the cycle formed by the extra edge must be of odd length. Let $L$ be the length of the cycle and w.l.o.g let its vertices be $\{1,\ldots,L\}$ with the corresponding framework points $p_1,\ldots,p_L$. The linear system (\ref{completion}) attached to the cycle edges is given by
\begin{equation}
\label{cycle-eqs}
p_i \dot{p}_{i+1} + p_{i+1}\dot{p}_i = 0,\; i=1,\ldots,L,
\end{equation}
(with the convention that $L+1$ is 1). The system (\ref{cycle-eqs}) results in the coefficient matrix
$$      C_L = \left(
      \begin{array}{cccccc}
      p_2 & p_1 & 0 & 0 & 0 & 0 \\
      0 & p_3 & p_2 & 0 & 0 & 0 \\
      0 & 0 & \ddots & \ddots & 0 & 0 \\
      0 & 0 & 0 & p_{L-1} & p_{L-2} & 0 \\
      0 & 0 & 0 & 0 & p_L & p_{L-1} \\
      p_L & 0 & 0 & 0 & 0 & p_1
      \end{array}
      \right)
$$
Expanding by the first column, we get that $$\det C_L = p_2 \prod_{i\neq 2} p_i + (-1)^{L+1}p_L \prod_{i\neq L}p_i = \left[1+(-1)^{L+1}\right] \prod_{i=1}^L p_i.$$
For $L$ even the determinant vanishes, from which it follows that there is a linear dependence (redundancy) among the equations and $G$ cannot be minimally locally completable. Therefore, the length of the cycle must be odd.

Conversely, if each connected component of $G$ is a tree plus one edge that forms an odd cycle, then from $\det C_L \neq 0$ it follows that the only solution to (\ref{cycle-eqs}) is the trivial one: $\dot{p}_1=\dot{p}_2=\ldots=\dot{p}_L=0$. From the connectivity of the connected component we get that all velocities must vanish, that is, the only solution to $C_G(p)\dot{p}=0$ is $\dot{p}=0$ and $G$ is locally completable. Since each connected component is a tree plus one edge it follows that $|E(G)|=|V(G)|$ which means that $G$ is also minimally locally completable.
\end{proof}

We now turn to the combinatorial characterization of global completion for Gram matrices in one dimension. Note that the inner products accessible to the analyst  are simply products of the form $p_i p_j = J_{ij}$ for $(i,j)\in E$. Taking the logarithm of the modulus gives the linear system
\begin{equation}
\label{log-abs}
q_i + q_j = \log |J_{ij}|, \quad (i,j)\in E,
\end{equation}
where $q_i=\log |p_i|$. This linear system has much resemblance with the linear system (\ref{completion}) of $C_G(p)\dot{p}=0$. Indeed, by substituting $p_1=p_2=\ldots=p_n=1$ in $C_G(p)$ the two coefficient matrices of (\ref{completion}) and (\ref{log-abs}) are the same. The only difference between to the systems is that (\ref{log-abs}) is non-homogenous. Therefore, by following the steps of the proof of Proposition \ref{prop:local-gram} it follows that if $G$ is minimally locally completable then the system (\ref{log-abs}) has a unique solution. That is, if each connected component of $G$ is a tree plus one edge that forms an odd cycle, then all the $|p_i|$'s are uniquely determined. This leaves only the signs of the $p_i's$ undetermined. Clearly, if $G$ has more than one connected component then the relative sign between coordinates of different components cannot be determined. Therefore, a globally completable graph must be connected. On the other hand, if $G$ is a single tree plus an edge that forms an odd cycle then by similar considerations all bit signs $b_i=(1-\operatorname{sign}p_i)/2\in \{0,1\}$ are determined up to overall negation (reflection), because they satisfy a similar linear system over $Z_2$
\begin{equation}
b_i \oplus b_j = (1-\operatorname{sign}(J_{ij}))/2,\quad (i,j)\in E.
\end{equation}
This is summarized in the following proposition:
\begin{proposition} \label{gram_global_d1}
  A graph is minimally Gram globally completable in one dimension if and only if it is a tree plus an edge that forms an odd cycle.
\end{proposition}

We now give the combinatorial characterization of local and global completion for the case of a general rectangular rank-1 matrices. The proofs are very similar to the Gram case and are therefore omitted. The the only differences being the number of degrees of freedom ($d^2$ instead of $d(d-1)/2$, which for $d=1$ are 1 and 0, respectively) and the fact that the underlying graph is bipartite.

\begin{proposition}
A rectangular graph is minimally locally completable in one dimension if and only if it is a forest (a disjoint union of trees, or equivalently, a $(1,1)$-tight sparse bipartite graph).
\end{proposition}

\begin{proposition}
A rectangular graph is minimally globally completable in one dimension if and only if it is a tree.
\end{proposition}

We note that the conditions for local and global completion in one dimension are much weaker than the condition for local and global rigidity in one dimension, which are connectedness and 2-connectedness, respectively. Laman theorem for two-dimensional rigidity leads us to speculate that the combinatorial characterization of Gram and rectangular rank-2 matrices will perhaps involve $(2,1)$-sparse graphs and bipartite $(2,4)$-sparse graphs. However, we currently postpone the investigation of this interesting question.

\section{Numerical Simulations}
\label{sec:numerical}

To illustrate the applicability of the above algorithms for testing low rank matrix completion in practice, we present the outcomes of several numerical experiments for both Gram and general rectangular rank-$d$ matrices.
Let us first investigate the case when $J$ is a Gram matrix of size $n \times n$. We apply the local completion Algorithm \ref{alg:local-gram} to test the local completion property of random graphs where each edge (including self loops) is chosen independently with probability $\beta$, such that the expected number of edges is $\mathbb{E} [m] = { n+1 \choose 2}\beta$. The conditional probability $f(n,d,\beta)$ for an $n\times n$ rank-$d$ Gram matrix to be locally completable
\begin{equation}
\label{cond-prob}
f(n,d,\beta) = \Pr\{G \mbox{ is Gram } n\times n \mbox{ rank-{\em d} locally completable}\,|\, \beta\}
\end{equation}
is a monotonic function of $\beta$. We define the threshold value $\beta^*(n,d)$ as the value of $\beta$ for which $f(n,d,\beta^*) = 1/2$. The theoretical bound of Cand\`es and Recht \cite{CandesCompletion} implies $\beta^*(n,d) \sim C(d) n^{-0.8}\log n$ (CR), while that of Keshavan, Oh, and Montanari \cite{Montanari1,Montanari2} gives $\beta^*(n,d) \sim C(d)n^{-1}\log n$ (KOM) for large $n$.

We now describe the numerical simulation and estimation procedure for the threshold values. The goal is to infer the asymptotic behavior of $\beta^*(n,d)$ for large values of $n$. For small values of $n$ the running time of the completion Algorithm \ref{alg:local-gram} is not an issue and it is possible to perform an exhaustive search for the threshold value. However, for large values of $n$ (e.g., $n=20000$) an exhaustive search is too time consuming.

The way we choose to accelerate the search is by using logistic regression. Suppose that after the $(i-1)$'th iteration we were able to estimate the threshold value for $n_{i-1} \times n_{i-1}$ matrices, and now we want to find the threshold for larger matrices of size $n_i \times n_i$ with $n_{i}>n_{i-1}$. From the previous threshold estimate $\beta^*(n_{i-1},d)$ we can easily guess a reasonable upper bound $\beta^u_i$ such that with probability one $\beta^*(n_{i},d) \leq \beta^u_i$ (e.g., $\beta^u_i = \beta^*(n_{i-1},d)$, because $\beta^*$ is decreasing with $n$). Starting from the upper bound $\beta=\beta^u_i$ we decrease $\beta$ in small steps until we encounter enough matrices that are not completable anymore (say until we observe 20 consecutive non-completable matrices). This gives us a rough confidence interval for $\beta^*$. We then sub-sample this interval to get a more accurate estimation of the threshold value by collecting more $\beta$'s and their corresponding binary responses $y\in \{0,1 \}$, where $y=1$ if the matrix is completable, and $y=0$ otherwise. We perform a binomial logistic regression in order to estimate numerically the approximate value of $\beta^{*}$. Logistic regression is a model frequently used in statistics for prediction of the probability of occurrence of an event, by fitting data to a logistic curve. The logistic function $f(\beta)$ models the conditional probability (\ref{cond-prob}) by
$$  \Pr\{y=1|\beta\} \approx  f(\beta) = \frac{1}{1+e^{-\alpha^*(\beta-\beta^{*})}},$$
where $\beta^{*}$ and $\alpha^{*}$ are parameters to be estimated from the set of samples of the explanatory variable $\beta$ and the corresponding binary responses $y$.
We typically used a total of about 120 $(\beta,y)$ sample pairs for the logistic regression (80 samples to find the lower bound, and another 40 samples to improve the confidence intervals for the estimate).
We performed similar searches and logistic regressions to find the threshold value for global completion of Gram matrices, as well as local and global completion of rectangular matrices.

Figure \ref{fig:gram} is a log-log plot of the threshold $\beta^{*}$ against $n$, for rank-$d$ Gram matrices (left) and rectangular matrices (right), with $d=2,3,4$. For ``rectangular" matrices we chose for simplicity $n_1=n_2=n$. The dotted curves that appear in green are the CR bounds, the blue dashed curves represent global completion, while the remaining three curves denote local completion.
\begin{figure}
\begin{center}
 \includegraphics[width=0.49\textwidth]{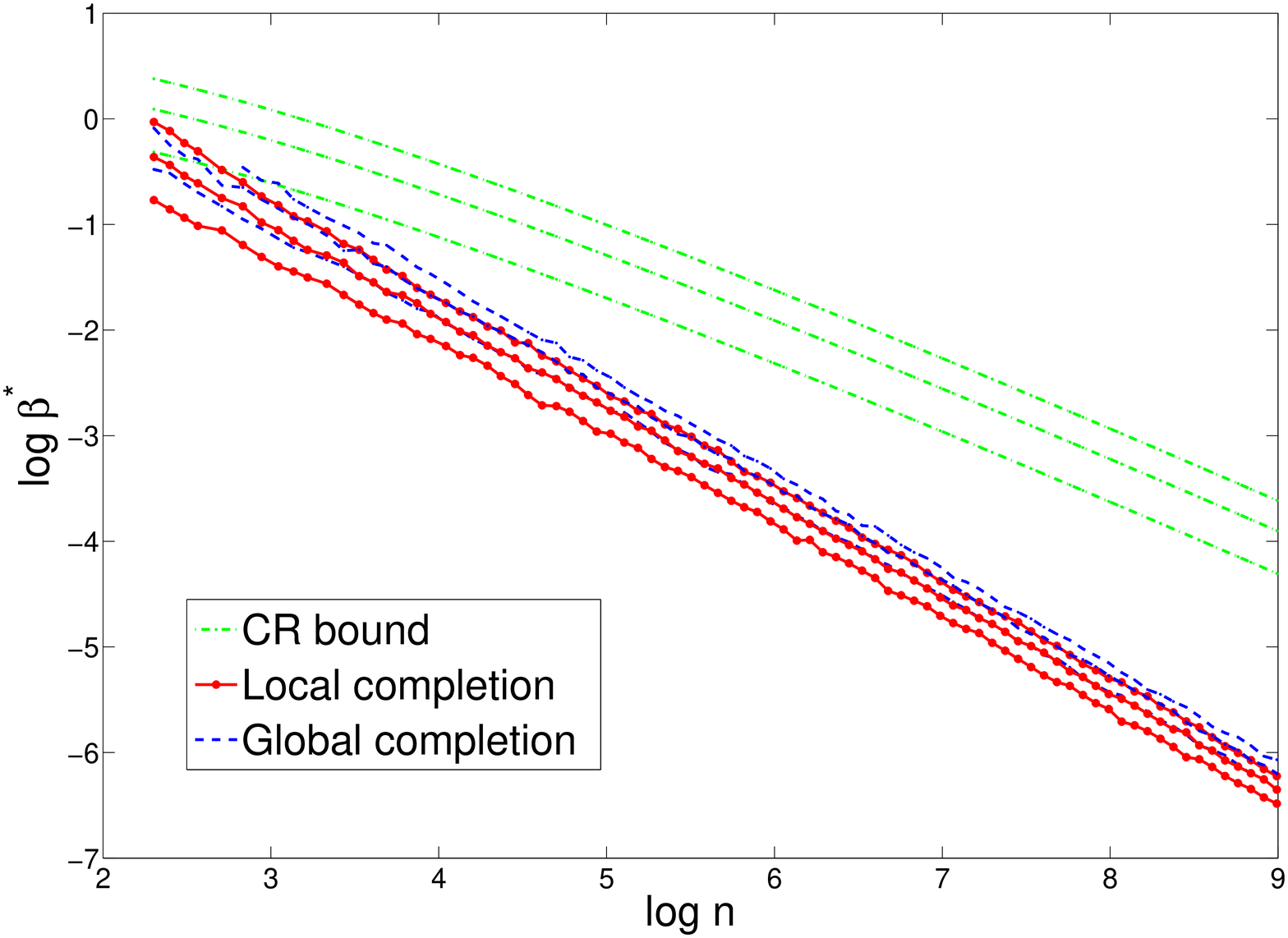}
 \includegraphics[width=0.49\textwidth]{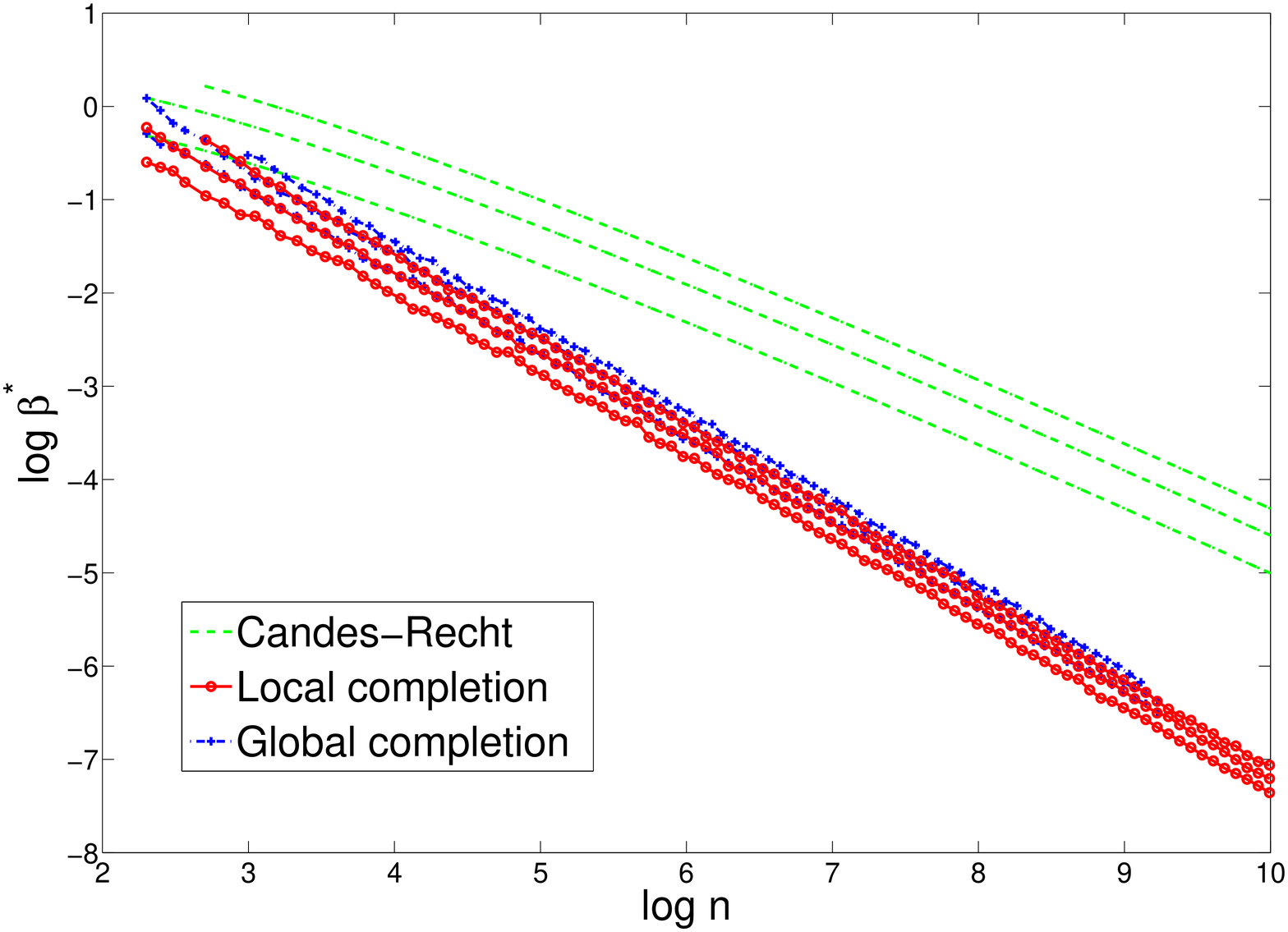}
\end{center}
\caption{Local and global completion of rank-$d$ Gram matrices (left) and rectangular matrices (right): the threshold $\beta^*$ as a function of $d$ and $n$ for $d=2,3,4$.} \label{fig:gram}
\end{figure}
It is interesting to note that both in the Gram case and in the rectangular case, the global completion curves for dimension $d$ coincide with the local completion curves for dimension $d+1$. This can perhaps be explained by the following theorem of Gortler, Healy, and Thurston \cite[Theorem 1.17]{Harvard1}: If a graph is generically locally rigid but not generically globally rigid in $\mathbb{R}^d$, then any generic framework can be connected to an incongruent framework by a path of frameworks in $\mathbb{R}^{d+1}$ with constant edge lengths. For example, the graph on 4 vertices with all possible edges except for one ($K_4$ minus an edge) is locally rigid in $\mathbb{R}^2$ but is not globally rigid in $\mathbb{R}^2$. This graph can be realized as two triangles with a common side and has two possible realizations in the plane. Though it is impossible to continuously deform the framework in the plane, it is possible to continuously deform it by leaving the plane into the three dimensional space by folding one triangle on top of the other. We may expect a similar phenomenon in the matrix completion problem: if a graph is not globally rank-$d$ completable, then it is not locally rank-$(d+1)$ completable.

Both the CR and the KOM bounds can be written as $\beta^* \sim C(d) n^\alpha \log n$ or upon taking the logarithm as $\log \beta^* \sim \alpha \log n + \log \log n + \log C(d)$.
We therefore performed a simple linear regression in the Gram case with $d=2$ of the form  $\log \beta^*(n) = a_1 \log n + a_2 \log \log n +a_3$ to estimate $a_1,a_2,a_3$ (see Table \ref{tab:reg}).
For local completion, the sampled $n$ take values up to 100000 while for global completion up to 15000, in a geometric progression of rate $1.08$. The coefficient $a_2$ may be expected to be 1 (the coupon collector problem) and Theorem \ref{Th-Jackson} for planar rigidity \cite{Jackson} may even shed light on the higher order corrections. The results for different rank values $d=2,3,4$ are summarized in Table \ref{tab:reg2}. These results may be considered as a numerical evidence for the success of the KOM theoretical bound $C(d)n\log n$. The slight deviation of $a_1+2$ from unity can be explained by the small bias introduced by the small $n$ values.
\begin{table}[h]
\begin{center}
\begin{tabular}{|l||l|l||l|l|}
\hline
 &\multicolumn{2}{l|}{Local Completion}&\multicolumn{2}{l|}{Global Completion}\\
\cline{2-5}
 &Value & 95\% -conf. interval & Value & 95\% conf. interval\\
\hline\hline
$a_1+2$ & 1.022   & [0.99842, 1.0456] & 0.99066 & [0.94206      1.0393]\\
$a_2$ & 0.63052 & [0.43626, 0.8247] & 0.79519 & [0.43446      1.1559]\\
$a_3$ & 0.90663 & [0.69405, 1.1192] & 0.99899 & [0.6394      1.3586] \\
\hline
$a_1+2$ & 0.9773   & [0.9748    0.9797] & 0.9631 & [0.95959     0.96668]\\
$a_2=1$ & - & - & - & - \\
$a_3$ & 0.5039 & [0.48295     0.5250] & 0.7954 & [0.76823     0.82258] \\
\hline
\end{tabular}\end{center}
\caption{Linear regression for local (left column) and global (right column) rank-2 Gram completion: $\log\beta^* = a_1 \log n + a_2 \log \log n +a_3$.}\label{tab:reg}
\end{table}
\begin{table}[h]
\begin{center}
\begin{tabular}{|l||l|l|}
\hline
   $d$   & ($a_1+2$,$a_2$,$a_3$) & ($a_1+2$,$1$,$a_3$) \\
\hline\hline
$2$ & (1.022, 0.63052 , 0.90663) & (0.9773, 1, 0.5039) \\
$3$ & (1.0423, 0.40053 , 1.394)   & (0.9636, 1, 0.78538)  \\
$4$ & (1.0382, 0.35039 , 1.6725)  & (0.95289, 1, 1.013)   \\
\hline
\end{tabular}\end{center}
\caption{Linear regression for local Gram completion for $d=2,3,4$.}\label{tab:reg2}
\end{table}

The asymptotic behavior $\beta^* \sim C(d) d n^{-1}\log n$ implies that $\frac{ \beta^{*} n}{ d \log n}$ tends to a constant as $n \to \infty$. Figure \ref{fig:constant} shows $\frac{\beta^*n}{d \log n}$ against $\log n$ for different values of $d$ in the Gram case (left) and in the rectangular matrix case (right) from which the asymptotic behavior is clear.

\begin{figure}
\begin{center}
 \includegraphics[width=0.49\textwidth]{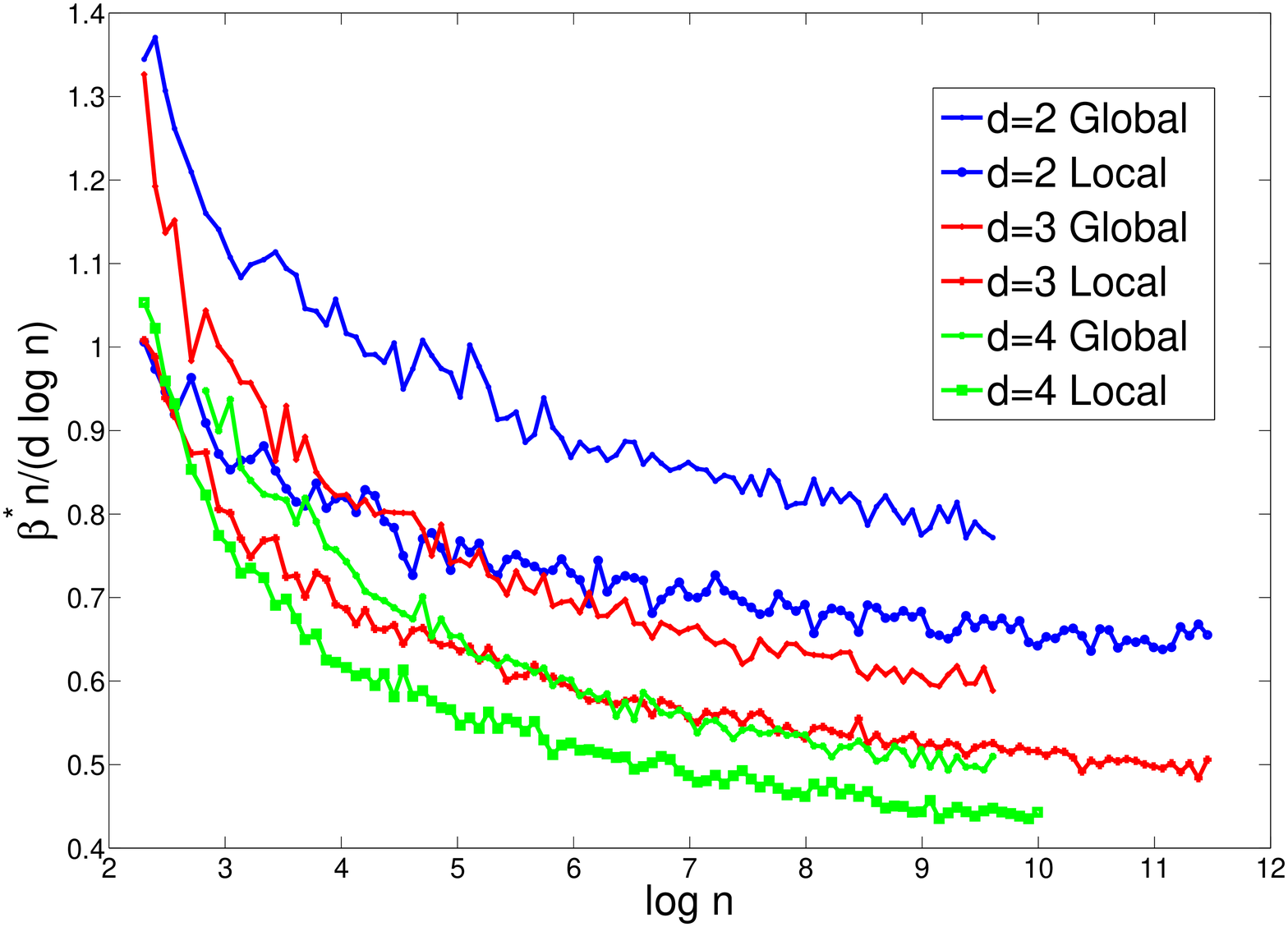}
 \includegraphics[width=0.49\textwidth]{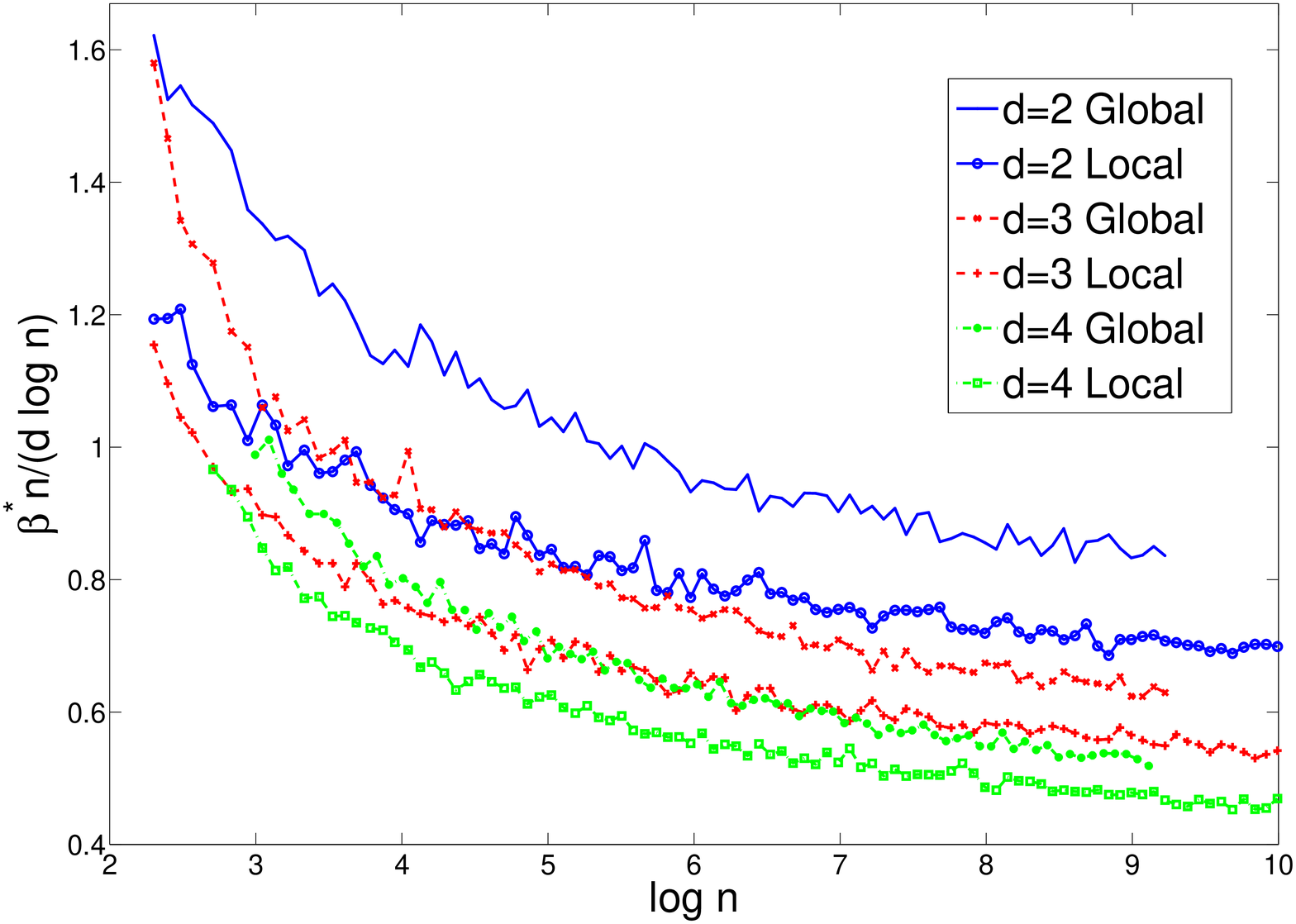}
\end{center}
\caption{A plot of $\frac{\beta^*n}{d \log n}$ vs. $\log n$ for Gram matrices (left) and rectangular matrices (right). For large values of $n$ we expect $\frac{\beta^*n}{d \log n}$ to go to the constant $C(d)$ of the KOM bound.}\label{fig:constant}
\end{figure}

Finally, in Tables \ref{tab:timelocal} and \ref{tab:timeglobal} we list the running times in seconds for several of the numerical simulations. All computations were performed on a PC machine equipped with an Intel(R) Core(TM)2 Duo CPU 3.16GHz with $4$ GB RAM. The matrices we considered were general ``rectangular" $n$-by-$n$ matrices of rank $2$. We list the running times for different values of $\beta = \frac{m}{n^2}$ as a function of $\beta^{*}$, where $m$ denotes the number of given entries in the matrix. The algorithm often runs faster as $m$ (alternatively, $\beta$) increases, because the iterative LSQR method converges after a fewer number of iterations. Note that the values $\beta=0.75 \beta^{*}$ in the first column in Table \ref{tab:timelocal}  produce matrices that are not locally completable. Similarly, in Table \ref{tab:timeglobal}, all values in the first column were chosen such that the matrix is locally completable, but not globally completable. In both tables, all the values $\beta=2\beta^{*}$ produce matrices that are locally (and globally) completable. We do this to better illustrate the time difference between the case when the LSQR method converges after a certain relatively small number of iterations, as opposed to when the residual cannot be made smaller than the threshold.

\begin{table}[h]
\begin{center}
\begin{tabular}{|l|  l|l|l||  l|l|l||  l|l|l || l|l|l| }
\hline
 & \multicolumn{3}{l||}{}  &\multicolumn{3}{l||}{ }  &\multicolumn{3}{l||}{} \\
$n$ & $\beta$ & $m$  & $T$  & $\beta$ & $m$  & $T$ & $\beta$ & $m$  & $T$ \\

\hline\hline
$101$  &  $0.75\beta^{*} $ & 598   & 0.07  & $\beta^{*}$ & 796 & 0.05 & $2\beta^{*}$ & 1592 & 0.06 \\
$498$  &  $0.75\beta^{*} $ & 3596  & 0.24  & $\beta^{*}$ & 4795 & 0.32 & $2\beta^{*}$ & 9590 & 0.25 \\
$1002$ &  $0.75\beta^{*}$  & 7792  & 0.75  & $\beta^{*}$ & 10390 & 0.85 & $2\beta^{*}$ & 20781 & 0.72 \\
$2533$ &  $0.75\beta^{*}$  & 21591 & 2.6   & $\beta^{*}$ & 28788 & 2.3 & $2\beta^{*}$ & 57575 & 2 \\
$5068$ &  $0.75\beta^{*}$  & 45982 & 10.9  & $\beta^{*}$ & 61309 & 7.3 & $2\beta^{*}$ & 122620 & 10.7 \\
$10133$&  $0.75\beta^{*}$  & 99140 & 28.6  & $\beta^{*}$ & 132190 & 12 & $2\beta^{*}$ & 264370 & 18.2 \\
\hline
\end{tabular}\end{center}
\caption{Running time $T$ (in seconds) in the case of \textit{local} completion of general $n$-by-$n$ rectangular matrices of rank $d=2$, for various values of $m$ (number of entries revealed). } \label{tab:timelocal}
\end{table}

\begin{table}[h]
\begin{center}
\begin{tabular}{|l| l|l|l||  l|l|l||  l|l|l || l|l|l| }
\hline
  &\multicolumn{3}{l||}{}  &\multicolumn{3}{l||}{ }  &\multicolumn{3}{l||}{} \\
$n$ &  $\beta$ & $m$  & $T$  & $\beta$ & $m$  & $T$ & $\beta$ & $m$  & $T$ \\
\hline\hline
$101$  &  $0.8 \beta^{*} $ & 800 & 0.17 &  $\beta^{*}$ & 999 & 0.14 &  $ 2\beta^{*}$ & 1998     & 0.14  \\
$498$  &  $0.8 \beta^{*} $ & 4638 & 1.3 &  $\beta^{*}$ &  5797 & 0.9 & $ 2\beta^{*}$ & 11595    & 0.88  \\
$1002$ &  $0.9 \beta^{*} $ & 11539 & 2.2 & $\beta^{*}$ & 12821 & 3.1 & $ 2\beta^{*}$ & 25642    & 2.3   \\
$2533$ &  $0.9 \beta^{*}$ & 31078 & 12 &   $\beta^{*}$ & 34531 & 10.1& $ 2\beta^{*}$ & 69063    & 16.1  \\
$5068$ &  $0.91\beta^{*}$ & 69020 & 45.2 & $\beta^{*}$ & 75847 & 43 &  $ 2\beta^{*}$ & 151690   & 59.4  \\
$10133$&  $0.92\beta^{*}$ & 143790 & 179 & $\beta^{*}$ & 156280 & 180 & 2$\beta^{*}$ & 312590 & 288   \\
\hline
\end{tabular}\end{center}
\caption{Running time $T$ (in seconds) in the case of \textit{global} completion of general $n$-by-$n$ rectangular matrices of rank $d=2$, for various values of $m$ (number of entries revealed). } \label{tab:timeglobal}
\end{table}

\section{Summary and Discussion}
\label{sec:summary}
In this paper we made the observation that the rank-$d$ matrix completion problem is tightly related to rigidity theory in $\mathbb{R}^d$, with inner products replacing the role of distances. Many of the results in rigidity theory, both classical and recent, give new insights into the completion problem. In particular, we introduced the completion matrix and stress matrices that enable fast determination of generic local and global completion properties of any partially viewed matrix into a rank-$d$ matrix. Our algorithms determine if a unique completion is possible without attempting to complete the entries of the matrix.

Most of the results in rigidity theory translate nicely into the completion setup. However, due to the large body of work in rigidity theory, as well as occasional differences between completion and rigidity, many of these results (such as generalization of Laman theorem to the completion setup) did not find their way into the current paper. Finally, we note that beyond its mathematical importance, rigidity theory is much useful in diverse applications such as scene analysis and localization of sensor networks, and some of the recent localization algorithms based on rigidity \cite{AmitPNAS,Harvard2} can be easily adjusted to the inner products completion setup.

\section*{Acknowledgments}
It is a pleasure to thank Haim Avron and Sivan Toledo for many valuable suggestions and interesting discussions on finding the null space of sparse matrices, sparse factorizations and iterative methods; Vladimir Rokhlin, Yoel Shkolnisky and Mark Tygert on similar aspects of numerical linear algebra; Aarti Singh for bringing to our attention recent developments in the matrix completion problem; Ronen Basri for discussions on the low rank matrix completion problem in the structure from motion problem in computer vision; and Gil Kalai and Paul Seymour for discussions on graph rigidity and sparse graphs.


\begin{thebibliography}{99}

\bibitem{Netflix}
ACM SIGKDD and Netflix. Proceedings of KDD Cup and Workshop, 2007. Proceedings available online
at {\tt http://www.cs.uic.edu/~liub/KDD-cup-2007/proceedings.html}.

\bibitem{Kanade}
Tomasi, C. and Kanade, T. (1992)
\newblock Shape and Motion from Image Streams under Orthography: a Factorization Method.
{\em International Journal of Computer Vision}, {\bf 9} (2):137--154.

\bibitem{SrebroPHD}
Srebro, N. (2004)
Learning with Matrix Factorizations.
{\em PhD thesis}, Massachusetts Institute of Technology (MIT).

\bibitem{Mesbani}
Mesbahi, M. and Papavassilopoulos, G.P (1997)
On the rank minimization problem over a positive semidefinite linear matrix inequality.
{\em IEEE Transactions on Automatic Control}, {\bf 42}(2):239--243.


\bibitem{Jacobs}
Jacobs, D. (1997)
Linear Fitting with Missing Data: Applications to Structure-from-Motion and to
Characterizing Intensity Images.
{\em IEEE Computer Society Conference on Computer Vision and Pattern Recognition (CVPR'97)}, pp.206.

\bibitem{Srebro2003}
{Srebro, N.} and {Jaakkola, T.} (2003)
\newblock Weighted Low-Rank Approximations.
\newblock In {\em Proceedings of the Twentieth International Conference on Machine Learning (ICML)}, 720--727.

\bibitem{Fazel1}
Fazel, M., Hindi, H., and Boyd, S. (2004)
Rank Minimization and Applications in System Theory.
{\em Proceedings of American Control Conference}, Boston, Massachusetts, pp. 3273--3278.

\bibitem{FazelPHD}
Fazel, M. (2002)
Matrix Rank Minimization with Applications. Elec. Eng. Dept, Stanford University.

\bibitem{CandesSVT}
Cai, J.-F., Cand\`es, E.J., and Shen, Z.
A Singular Value Thresholding Algorithm for Matrix Completion, {\em submitted for publication}.


\bibitem{SDP}
Vandenberghe, L. and Boyd, S.P. (1996)
Semidefinite programming.
{\em SIAM Review}, {\bf 38}(1):49--95.

\bibitem{CandesTao1}
Cand\`es, E.J. and Tao, T. (2005)
Decoding by linear programming.
{\em IEEE Transactions on Information Theory}, {\bf 51}(12):4203--4215.

\bibitem{Donoho}
Donoho, D.L. (2006)
Compressed sensing.
{\em IEEE Transactions on Information Theory}, {\bf 52}(4):1289--1306.

\bibitem{CandesCompletion}
Cand\`es, E.J. and Recht, B.
Exact Matrix Completion via Convex Optimization, {\em submitted for publication}.


\bibitem{Montanari1}
Keshavan, R., Oh, S., and Montanari, A. (2009)
Matrix Completion from a Few Entries. {\em ISIT 2009}, submitted.

\bibitem{Montanari2}
Keshavan, R., Montanari, A., and Oh, S. (2008)
Learning low rank matrices from $O(n)$ entries. {\em Allerton 2008}.

\bibitem{Roth1981}
{Roth, B.} (1981)
\newblock Rigid and Flexible Frameworks.
\newblock {\em The American Mathematical Monthly}, {\bf 88}(1):6--21.

\bibitem{Harvard1}
Gortler, S.J., Healy, A., and Thurston, D. (submitted)
Characterizing Generic Global Rigidity, {\em arXiv}:0710.0926.

\bibitem{Connelly}
Connelly, R. (2005) Generic global rigidity, {\em Discrete Comput. Geom} {\bf 33} (4), 549--563.

\bibitem{Connelly2}
Connelly, R. and Whiteley, W.J. (submitted) Global Rigidity: The effect of coning.


\bibitem{Hendrickson1992}
{Hendrickson, B.} (1992)
\newblock Conditions for unique graph realizations.
\newblock {\em SIAM J. Comput.} {\bf 21} (1):65--84.


\bibitem{Gluck}
{Gluck, H.} (1975)
\newblock Almost all simply connected closed surfaces are rigid.
\newblock In Geometric Topology, Lecture
Notes in Mathematics No. 438, Springer-Verlag, Berlin, pp. 225--239.


\bibitem{Asimow}
Asimow, L. and Roth, B. (1978) The rigidity of graphs. {\em Trans. Amer. Math. Soc.} {\bf 245}, 279-?289.

\bibitem{Maxwell}
Maxwell, J.C. (1864) On the calculation of the equilibrium
and stiffness of frames. {\em Philos. Mag.}, {\bf 27}:294?299.

\bibitem{Laman}
Laman, G. (1970) On graphs and rigidity of plane skeletal
structures. {\em Journal of Engineering Mathematics}, {\bf 4}:331?340.

\bibitem{Pebble1997}
{Jacobs, D. J.}  and {Hendrickson, B.} (1997)
\newblock An Algorithm for Two-Dimensional Rigidity Percolation: The Pebble Game.
\newblock {\em Journal of Computational Physics}, {\bf 137}:346-?365.

\bibitem{Lee}
Lee, A. and Streinu, I. (2008)
Pebble Game Algorithms and Sparse Graphs.
{\em Discrete Mathematics}
{\bf 308} (8), pp 1425--1437.


\bibitem{Connelly3}
Connelly, R. (1991) On generic global rigidity. In {\em Applied Geometry and
Discrete Mathematics}, DIMACS Ser. Discrete Math, Theoret. Comput.
Scie 4, AMS, pp. 147--155.

\bibitem{JacksonJordan}
Jackson, B. and Jordan, T. (2005) Connected rigidity matroids and
unique realization graphs. {\em J. Combinatorial Theory B} {\bf 94}, pp 1--29.

\bibitem{Jackson}
Jackson, B., Servatius, B., and Servatius, H. (2006) The 2-Dimensional Rigidity of Certain Families of Graphs. {\em Journal of Graph Theory} {\bf 54} (2), pp. 154--166.

\bibitem{Theran}
L. Theran (2008) Rigid components of random graphs. Arxiv preprint arXiv:0812.0872.

\bibitem{GotsmanToledo}
Gotsman, C. and Toledo, S. (2008)
On the Computation of Null Spaces of Sparse Rectangular Matrices.
{\em SIAM Journal on Matrix Analysis and Applications} {\bf 30} (2), pp. 445--463.

\bibitem{TimDavis}
Davis, T. (submitted) Multifrontal multithreaded rank-revealing sparse
QR factorization, submitted to {\em ACM Trans. on Mathematical Software}.

\bibitem{Golub}
Golub, G.H. and Van Loan, C.F. (1996)
{\em Matrix computations.}, 3rd ed., Johns Hopkins University Press, Baltimore, p.~694.


\bibitem{LSQR}
Paige, C.C. and Saunders, M.A. (1982)
LSQR: An Algorithm for Sparse Linear Equations and Sparse Least Squares.
{\em ACM Transactions on Mathematical Software (TOMS)} {\bf 8} (1):43--71.

\bibitem{AmitPNAS}
Singer, A. (2008)
A Remark on Global Positioning from Local Distances.
{\em Proceedings of the National Academy of Sciences}, {\bf 105} (28):9507--9511.

\bibitem{Harvard2}
Zhang, L., Liu, L., Gotsman, C., and Gortler, S.J. (2009)
An As-Rigid-As-Possible Approach to Sensor Network Localization.
{\em Harvard Computer Science Technical Report}: TR-01-09.


%
%
%
%
%
%
%


\end{thebibliography}
\end{document}